\documentclass{article}

\usepackage{microtype}
\usepackage{graphicx}
\usepackage{subfigure}
\usepackage{booktabs} 

\usepackage{hyperref}

\usepackage{graphicx} 
\usepackage{amsmath,amsthm,amssymb,bbm,stmaryrd,bm}
\usepackage{url}
\usepackage{dsfont}
\usepackage{color}
\usepackage{wrapfig}
\usepackage{listing}
\usepackage{amsopn}
\usepackage{algorithm}
\usepackage{algorithmic}

\usepackage{amsthm}

\def\abovestrut#1{\rule[0in]{0in}{#1}\ignorespaces}
\def\belowstrut#1{\rule[-#1]{0in}{#1}\ignorespaces}

\usepackage{hyperref}
\usepackage{url}
\usepackage{capt-of}
\usepackage{tabu}
\usepackage[usenames,dvipsnames,svgnames,table,x11names]{xcolor}

\usepackage{amsmath}
\newcommand{\yg}[1]{}
\newcommand{\dk}[1]{}
\newcommand{\dz}[1]{}

\usepackage[accepted]{icml2018}
\usepackage{natbib}
\usepackage{breqn}
\usepackage{collcell}
\usepackage{array}
\usepackage{tikz}
\usepackage{pgfkeys}
\usepackage{graphicx}

\pgfkeys{/heat/.is family, /heat,
  Max colour/.initial = Green4,
  Min colour/.initial = Red1,
  max colour/.initial = SpringGreen3,
  min colour/.initial = Yellow1,
  text colour/.initial = black,
  Min color/.style = {Min colour=#1},
  Max color/.style = {Max colour=#1},
  min color/.style = {min colour=#1},
  max color/.style = {max colour=#1},
  text color/.style = {text colour=#1},
  min/.initial = -1,
  max/.initial = 1,
  slider/.code={%
     \tikz{\shade[left color=\HVal{min colour},%
                  right color=\HVal{max colour}]%
        (current page.south west) rectangle ++(#1,12pt);
     }%
  }%
}
\newcommand\Heatset[1]{\pgfkeys{/heat, #1}}
\newcommand\HVal[1]{\pgfkeysvalueof{/heat/#1}}

\newcolumntype{H}{>{\collectcell\Heat}r<{\endcollectcell}}
\newcommand\Heat[1]{
\if\relax\detokenize{#1}\relax
\else%
  \pgfmathparse{int(100*(#1-\HVal{min})/(\HVal{max}-\HVal{min}))}
  \ifnum\pgfmathresult>100
    \edef\HeatCell{\noexpand\cellcolor{\HVal{Max colour}}}%
  \else\ifnum\pgfmathresult<0
      \edef\HeatCell{\noexpand\cellcolor{\HVal{Min colour}}}%
    \else
      \edef\HeatCell{\noexpand\cellcolor{\HVal{max colour}!\pgfmathresult!\HVal{min colour}}}%
    \fi%
  \fi%
  \HeatCell\textcolor{\HVal{text colour}}{$#1$}%
\fi%
}

\title{}

\begin{document}

\twocolumn[
\icmltitle{LaVAN: Localized and Visible Adversarial Noise}

\icmlsetsymbol{equal}{*}

\begin{icmlauthorlist}
\icmlauthor{Danny Karmon}{bar}
\icmlauthor{Daniel Zoran}{goo}
\icmlauthor{Yoav Goldberg}{bar}
\end{icmlauthorlist}

\icmlaffiliation{bar}{Department of Computer Science, 
Bar-Ilan University,Ramat Gan, Israel.}
\icmlaffiliation{goo}{DeepMind, London, UK.}

\icmlcorrespondingauthor{Danny Karmon}{karmond@biu.ac.il}
\icmlcorrespondingauthor{Yoav Goldberg}{yogo@cs.biu.ac.il}
\icmlcorrespondingauthor{Daniel Zoran}{danielzoran@google.com}

\icmlkeywords{Adversarial Examples, Deep Learning, Machine Learning, Vision}

\vskip 0.3in
]
\printAffiliationsAndNotice{}
\begin{abstract}

Most works on adversarial examples for deep-learning based image classifiers use
noise that, while small, covers the entire image. We explore the case where the
noise is allowed to be visible but confined to a small, localized patch of the
image, without covering any of the main object(s) in the image. We show that it 
is possible to generate localized adversarial noises that cover only 2\% of the pixels in the 
image, none of them over the main object, and that are transferable across
images and locations, and successfully fool a state-of-the-art Inception v3
model with very high success rates.

\end{abstract}
\begin{center}
\begin{tabular}{cc}
    \hbox{\includegraphics[scale=0.3]{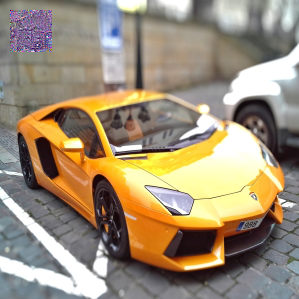}} &
    \hbox{\includegraphics[scale=0.3]{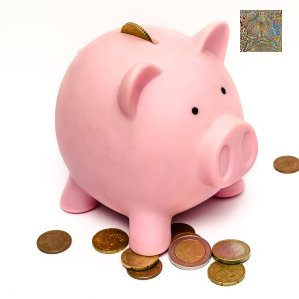}}\\
     Padlock (92.7\%) & Tiger Cat (94.4\%)\\
    \hbox{\includegraphics[scale=0.3]{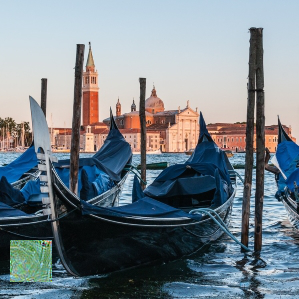}} &
    \hbox{\includegraphics[scale=0.3]{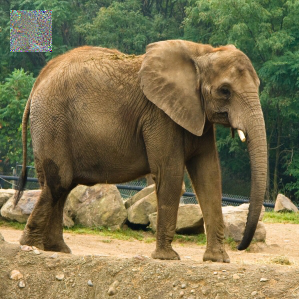}} \\
    Car Mirror (94.5\%) & Stingray (90.5\%) \\
\end{tabular}
\end{center}

\section{Adversarial Noise}

Deep neural-network architectures achieve remarkable results on image classification tasks. However, they are also susceptible to being fooled by adversarial examples: input instances which were modified in a particular way, and as a result, are misclassified by the network. Of course, for the adversarial example to be interesting, the change should be such that it does not confuse a human looking at the picture. Beyond the clear security implications, adversarial examples are also interesting as they may provide insights into the strengths, weaknesses, and blind-spots of these ubiquitous state-of-the-art classification models.

\begin{figure*}[ht]
\small
\begin{center}
\begin{tabular}{cc}
  \includegraphics[width=0.45\textwidth]{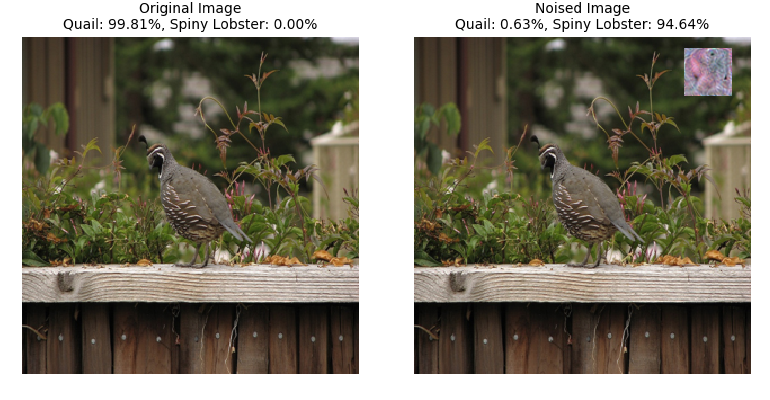} &\includegraphics[width=0.45\textwidth]{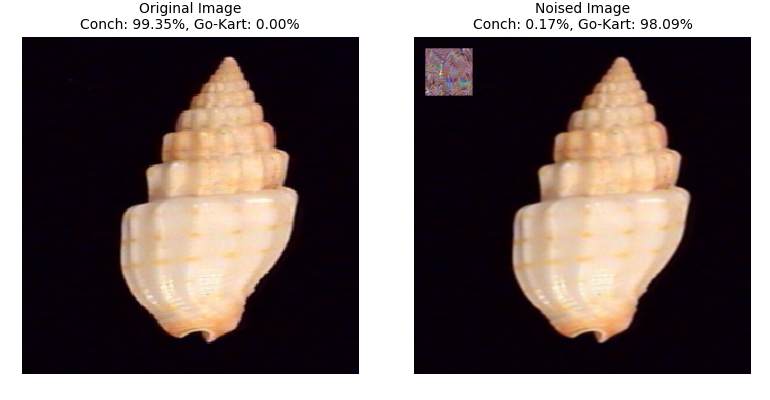}\\
  Quail (99.8\%) $\rightarrow$ Spiny Lobster (94.6\%) & Conch (99.4\%) $\rightarrow$ Go-Kart (98.1\%)\\[1em]
  \includegraphics[scale=0.39]{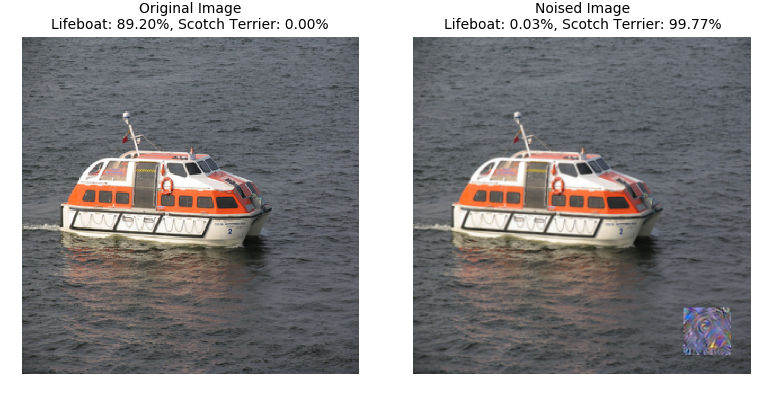}
  &\includegraphics[scale=0.39]{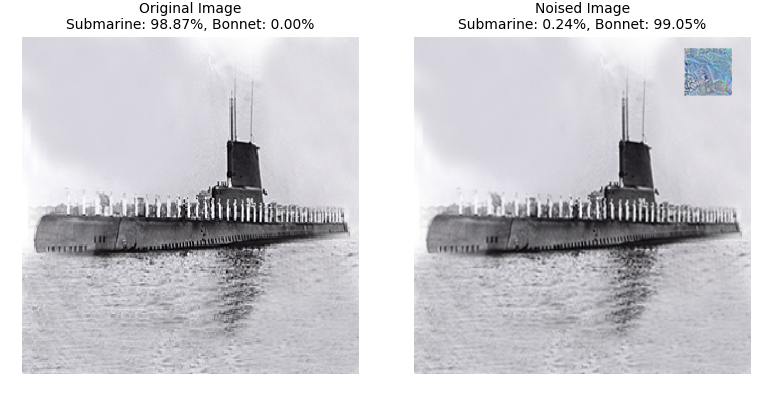}\\
   Lifeboat (89.2\%) $\rightarrow$ Scotch Terrier  (99.8\%)& Submarine (98.9\%) $\rightarrow$ Bonnet (99.1\%) \\
\end{tabular}
\end{center}

\caption{Images with network-domain localized noise. The Noise was generated for a specific input and location. Less than 2\% of pixels are noised. Network-domain noises are scaled to image-domain for presentation.}
\label{fig:single-image-noised}
\end{figure*}

Most work on generating adversarial examples (we provide a more detailed review in section \ref{sec:related}) focus either on noise which---while being imperceptible to humans---covers the entire image \cite{goodfellow2014explaining,szegedy2014intriguing}, or on visible noise that covers prominent features of the main object in the image in a ``natural'' way (i.e., glasses with a specific pattern around a person's eyes in a face identification task \cite{sharif2016accessorize}). 
In contrast, we look at \textbf{visible noise} that is \textbf{localized} to a
small area of the image (a bounded box with up to 2\% of the pixels), and which
\textbf{does not cover} the main object in the image. Figure \ref{fig:single-image-noised}
shows examples of such noised images that are misclassified by a state-of-the-art Inception V3 network with very high confidence.

A recent work by Brown et al \cite{brown2017adversarial} introduces a visible noise similar to ours. The works are complementary to a large extent.
Their work focuses on the security implications and attempts to generate
universal noise ``patches'' that can be physically printed and put on any image, in either a black-box
(when the attacked network is unknown) or white-box (when the attacked network
is known) setup. As a consequence, the resulting adversarial patches in \cite{brown2017adversarial} are relatively large (in
a white-box setup, the generated noise has to cover about 10\% of the image to be
effective in about 90\% of the tested conditions, and a disguised patch has to cover about 35\% of the image for a similar result) and also visually resemble the target class to some extent.
We do not attempt to produce a physical attack and are more interested in investigating the blind-spots of state-of-the-art image classifiers, and the kinds of noise that can cause them to misclassify.

We experiment with two setups: working in the \emph{Network Domain} and in the \emph{Image Domain}. In the network-domain case, the noise is allowed to take any value and is not restricted to the dynamic range of images. Such noise is akin to shining a very bright flash-light into someone's eye. In the image-domain case the noise is kept to the dynamic-range of images.

We show that in a white-box setting, we can generate localized visible noise
that can be transferred to almost arbitrary images, covers only up to 2\% of the image,  does not cover any part of the main object in the image and yet manages to make the network misclassify with very high confidence.  This works both for the network-domain and the image-domain case, although the success rates are naturally higher for the network-domain.

Moreover, by inspecting the gradients of the network over noised images, we show that the network does not usually identify the noised patch as the main cause of misclassification, and in some cases hardly assigns it any blame at all. The latter is true even in network-domain case. This is in contrast to the hypothesis posed in \cite{brown2017adversarial}, in which the noise is said to be ``much more salient'' to the neural network than real-world objects. 

The localized noises we generate are universal in the sense that they can be applied to many different images and locations. However, they are specific to a model they were trained on (i.e., equivalent to the white-box setups in \cite{brown2017adversarial}).

We believe these results highlight an interesting blind-spot in current state-of-the-art network architectures.

\section{Localized noise for a single image and location}
In the first setup, we explore generating a visible but localized adversarial
noise that is specific to a \emph{single image} and \emph{location} within this image. 

\subsection{Setting and Method}
Our method mostly follows that standard adversarial noise generation setup: we assume access to a trained model $M$ that assigns membership probabilities $p_M(y | x)$ to input images $x \in R^{n = w \times h \times c}$. We denote by $\Vec{y} = p_M(x)$ the vector of all class probabilities, and by $y=\arg\max_{y'} p_M(y=y'|x)$ the highest scoring class for input $x$ (the classifier's prediction). Let $y_{\text{source}}$ be the classifier's prediction on input $x$ (the \emph{source class}). We seek an image $x'$ that is classified by the network as $y_{\text{target}}$ (the \emph{target class}). The image $x'$ is composed of the original image with an additive noise $\delta\in R^n$: $x' = x+\delta$.

This is cast an optimization problem, seeking a value $\delta$ to to maximize $p_M(y=y_{\text{target}} | x+\delta)$.
The noise $\delta$ can be found using a stochastic gradient based algorithm.

We depart from this standard methodology by:
\begin{enumerate}
\item We want the noise $\delta$ to be confined to a small area over the image $x$, and to replace this area rather than be added to it. This is achieved by setting a mask $m\in \{0,1\}^n$, and taking the noised image to be $(1-m)\odot x-m\odot \delta$, where $\odot$ is element-wise multiplication.

\item Instead of training the noise to either maximize the probability of the target
class or to minimize the probability of any other class (including the source class), we use a loss that does both things---it attempts to move the prediction towards the target class and away from the highest scored class. We use the network activations prior to the final softmax layer, denoted as $M(x)$, $M(y=y'|x)$. This decouples the outputs for the different classes, and speeds up convergence.

\begin{dmath}
  \arg\max_\delta \left[ M(y=y_{\text{target}} | (1-m)\odot x+m\odot \delta) - M(y=y_{\text{source}} | (1-m)\odot x+m\odot \delta)\right]
\end{dmath}


\end{enumerate}

\paragraph{Network vs Image domain}
Pixels values in each channel can take byte values (0-255), which are scaled floats in the range [0,1] when used as input to the network. We experiment with two noising setups: \emph{network domain} and \emph{image domain}. In the \emph{network domain} case, the noise values is not restricted, and can go beyond the [0,1] dynamic range of the network. In the \emph{image domain} case, the noise is restricted to be within 0 and 1. To facilitate image domain noises, we clip the noise patch to the range [0,1] after each gradient step.\yg{should we say more or does this suffice?}\dk{I think that's good enough}

The entire process is detailed in algorithm \ref{alg:alg1}.

\begin{algorithm}[b!]
   \caption{Localized Noising Process}
   \label{alg:alg1}
\begin{algorithmic}
   \STATE {\bfseries Input:} image $x$, model $p_M$, target class $y_{\text{target}}$, target probability $s$, mask $m$, noise domain $d$.
   \STATE $y_{\text{source}} = \arg\max_y p_M(y | x)$
   \STATE $\delta = \Vec{0}$ 
   \STATE $x' = (1-m)\odot x+m\odot \text{noise}$
   \REPEAT 
   \STATE $\Vec{y} = p_M(x')$
   \STATE $L_{\text{target}} = \ell(\Vec{y}, y_{\text{target}}) = M(y=y_{\text{target}} | x')$ 
   \STATE $L_{\text{argmax}} = \ell(\Vec{y}, y_{\text{argmax}}) = M(y=y_{\text{argmax}} | x')$ 
   
   \STATE $\nabla_{\text{target}} = \frac{\partial L_{\text{target}}}{\partial x}$ 
   \STATE $\nabla_{\text{argmax}} = \frac{\partial L_{\text{argmax}}}{\partial x}$ 
   \STATE $\delta = \delta - \epsilon\cdot(\nabla_{\text{target}} - \nabla_{\text{argmax}})$
   \IF{$d=$\textit{image domain}}
    \STATE $\delta = clip(\delta, 0, 1)$
    \ENDIF
    \STATE $x' = (1-m)\odot x+m\odot \delta$
   \UNTIL{$p_M(y=y_{target} | x') \geq s$}
\end{algorithmic}
\end{algorithm}

\begin{figure*}[t]
\begin{center}
    
\scriptsize
\begin{tabu}{ccc}
  \includegraphics[scale=0.27]{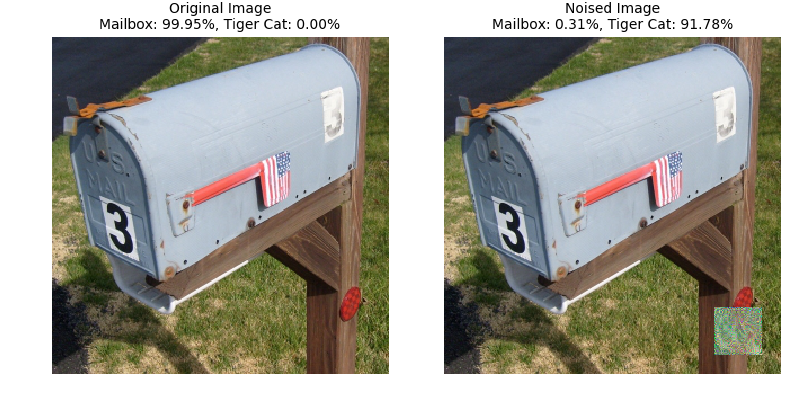} &\includegraphics[scale=0.27]{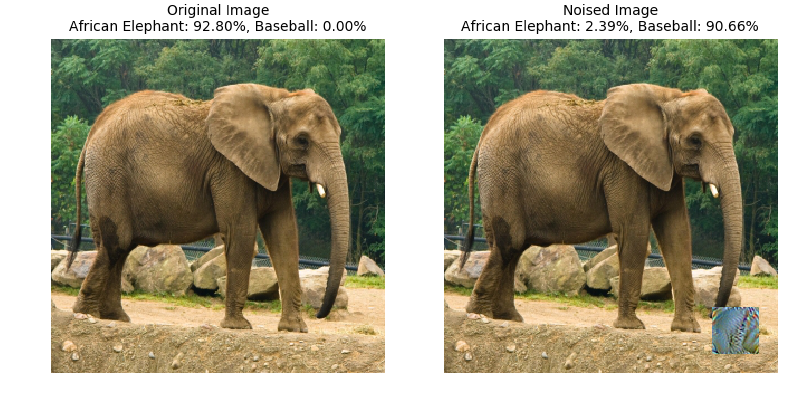}&\includegraphics[scale=0.27]{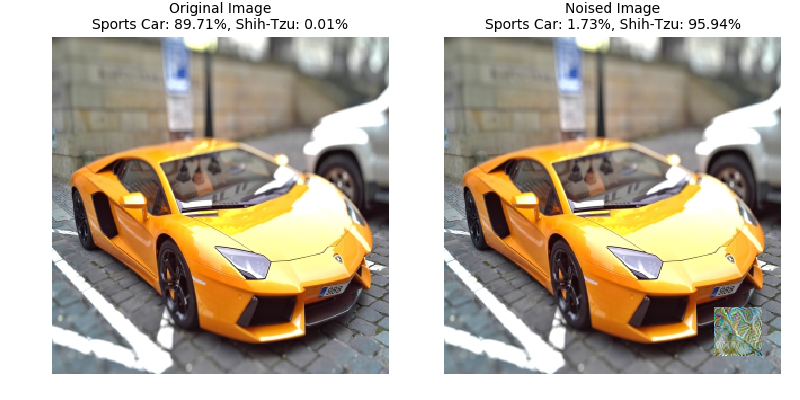} \\
  Mailbox (99.9\%) $\rightarrow$ Tiger Cat (91.8\%) & African-Elephant (92.8\%) $\rightarrow$ Baseball (90.7\%) & Sports Car (92.8\%) $\rightarrow$ Shih-Tzu (90.7\%)\\
  \\
  
  \includegraphics[scale=0.27]{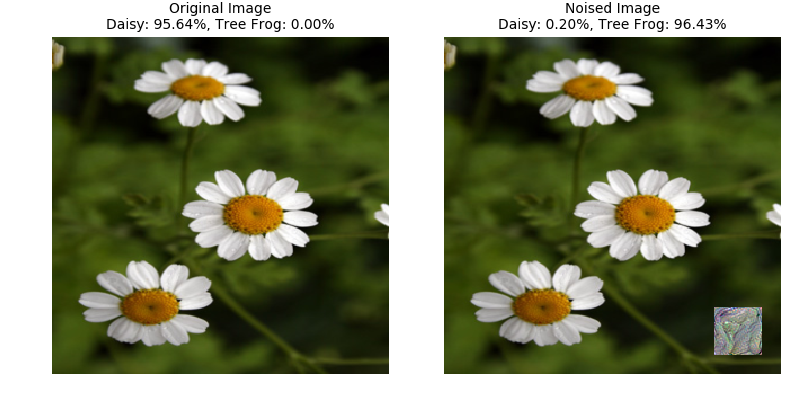} &\includegraphics[scale=0.27]{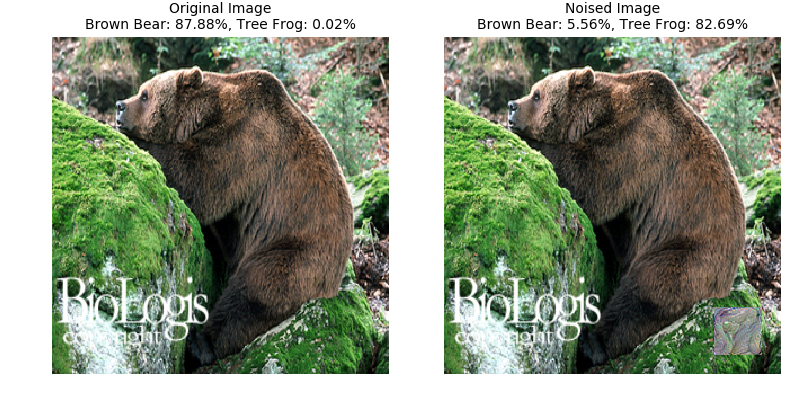}  &\includegraphics[scale=0.27]{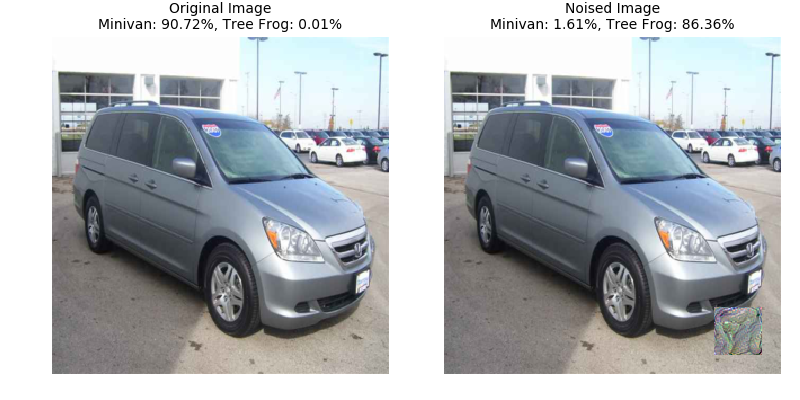} \\
  Daisy (95.6\%) $\rightarrow$ \textbf{Tree Frog} (96.4\%) & Brown Bear (87.9\%) $\rightarrow$ \textbf{Tree Frog} (82.7\%) & Minivan (90.7\%) $\rightarrow$ \textbf{Tree Frog} (86.4\%) \\
\end{tabu}

\caption{Transferable localized noises (network domain). Different network-domain noise patches, any of them can work on different images and locations. Notice that the same patch was used for all three images in the second row. The noises are scaled to image domain for visualization purposes.}
\label{fig:transferable-noised}
\end{center}
\end{figure*}

\begin{figure*}[t!]
 \centering
 \tiny
    \begin{tabu}{ccc}

 \includegraphics[scale=0.27]{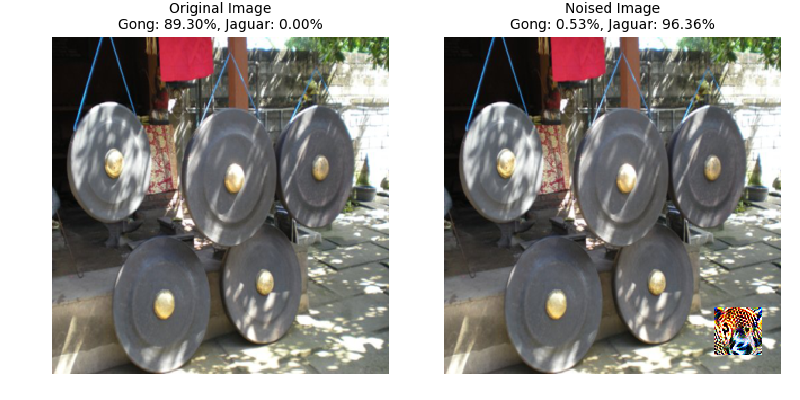} & \includegraphics[scale=0.27]{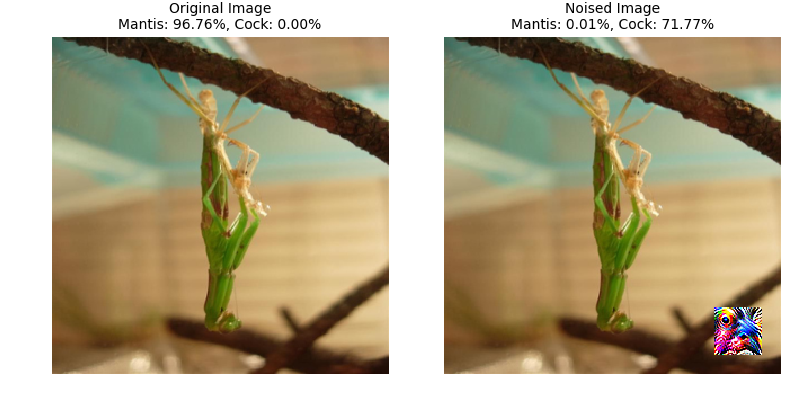} &
 \includegraphics[scale=0.27]{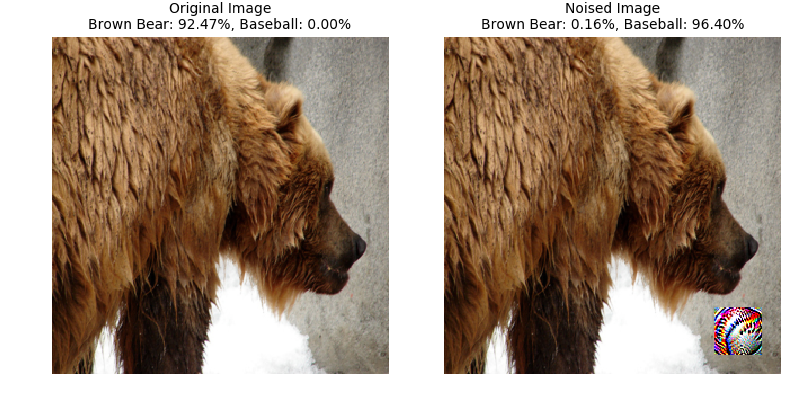}\\
   Gong (89.3\%) $\rightarrow$ \textbf{Jaguar} (96.4\%) & Mantis (96.8\%) $\rightarrow$ \textbf{Cock} (71.8\%) & Brown Bear (92.5\%) $\rightarrow$ \textbf{Baseball} (96.4\%) \\

 \end{tabu}
    \caption{Transferable localized noises (image domain).}
    \label{fig:transferable_image_domain}
\end{figure*}

\subsection{Experiments and Results}

We use the PyTorch provided pre-trained Inception V3 network \cite{szegedy2016rethinking} trained on ImageNet. We use images of size 299x299 and noise a square patch of size 42x42, roughly 2\% of the image pixels.  We generate network-domain noise this way for 100 random (image, target-class, location) triplets. The location is chosen around the corners (we focus on the corners as they have the
most chance of not covering the main object of the image). We consider the
noising process successful if the network classifies the noised image to the
desired target class with a confidence of at least 90\%.
For each image, we train the noise until we reach the desired confidence, or up to 10,000 iterations.

We managed to successfully generate localized network-domain noise this way to 79\% of the 110
configurations we tried. When relaxing the requirement such that the network
chooses the target class as the argmax (but perhaps with less than 90\% probability), we
manage to noise 91\% of the configurations. By further relaxing the criteria to
classify to any class other than the source one, noising success reaches 98\%. Figure \ref{fig:single-image-noised} shows examples of such noised images. Because network-domain noises are outside of the the range of valid pixels, we normalize the displayed noise patch to image range for visualization purposes.
 
\subsection{Shortcomings}
While we managed to produce a high confidence, targeted localized noise to 75 out of 100
(image, target-class, location) triplets we tried, the resulting noise is highly
dependent on the exact image and location: attempts to shift the noise even a single
pixel to either direction results in the network classifying to the original class
with high confidence. Similarly, attempts to move the noise to a different image
fail, unless we transfer, together with the noise, also a border of about 25\%
pixels to each side of the source image. In the next section, we generate noise
that can be moved across images and locations.

\section{Transferable localized noise}

We now explore transferable (``universal'') localized noises: a noise patch which can be applied across different images and image positions.

\subsection{Method}

We extend the localized noising process in Algorithm \ref{alg:alg1} by choosing, at each iteration, a random image $x$ from a ``training set'' of $100$ images, and a random location. We adjust the noise vector and the mask so that they correspond to the target location, apply the noise to the image, and take a gradient step over the noise away from the source class of $x$ and towards the shared target class $y_{\text{target}}$. Thus, at each iteration the same noise is applied on a random image and location, both sampled separately from uniform distributions of the possible images and locations respectively. This is very similar to the algorithm presented in \cite{brown2017adversarial}, with a somewhat different loss function, as described above. We stop the noise generation process after the prediction model misclassifies with the desired confidence (i.e. Target class probability $\geq 0.9$) for 30 consecutive iterations. 

\subsection{Experiments and Results}

\begin{figure*}[b!th]
    \centering
    \tiny
    \setlength\tabcolsep{0.5pt}
    \begin{tabular}{ccccccccccc}
      \textbf{Network-Domain} & \textbf{Image-Domain} &  & \textbf{Network-Domain} & \textbf{Image-Domain} & & \textbf{Network-Domain} & \textbf{Image-Domain} &  & \textbf{Network-Domain} & \textbf{Image-Domain} \\
        \hbox{\includegraphics[scale=0.14]{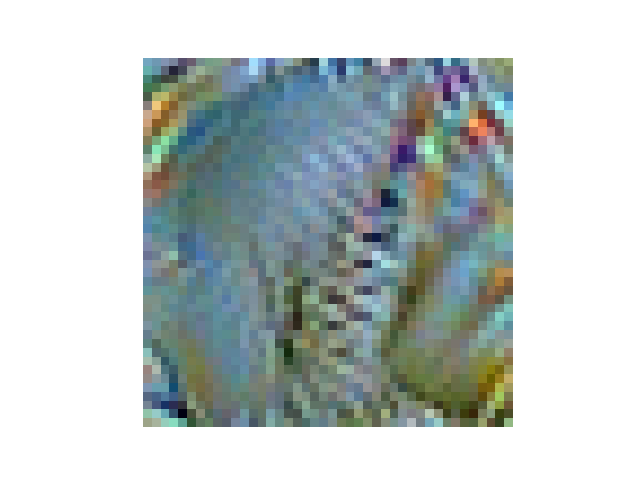}} &
        \hbox{\includegraphics[scale=0.14]{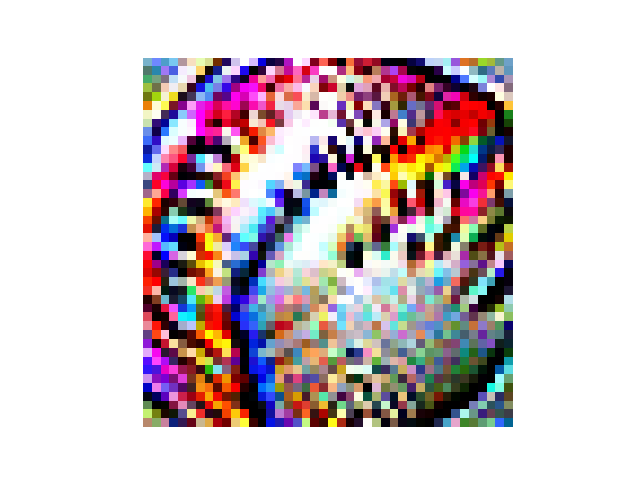}} &
        &
        \hbox{\includegraphics[scale=0.14]{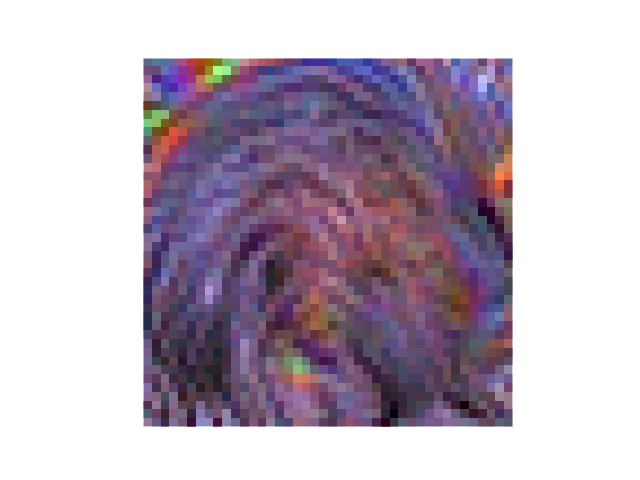}} &
        \hbox{\includegraphics[scale=0.14]{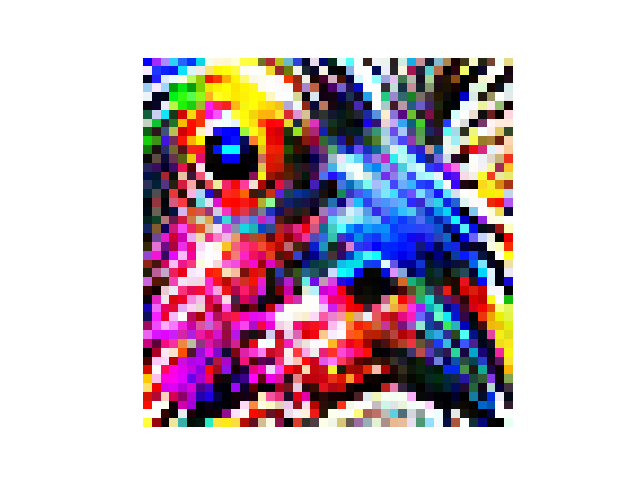}} &
        &
        \hbox{\includegraphics[scale=0.14]{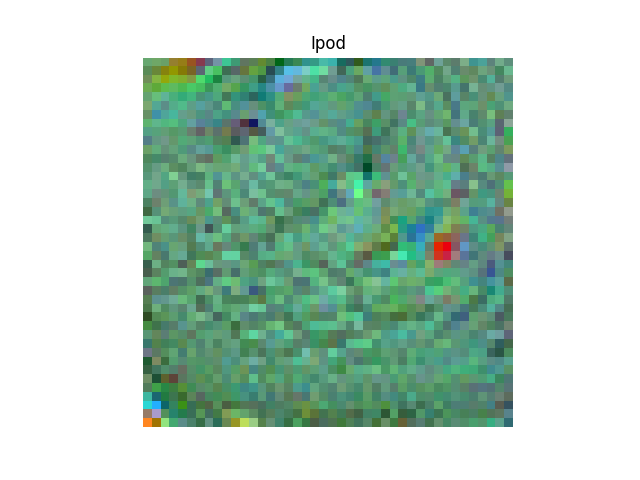}} &
        \hbox{\includegraphics[scale=0.14]{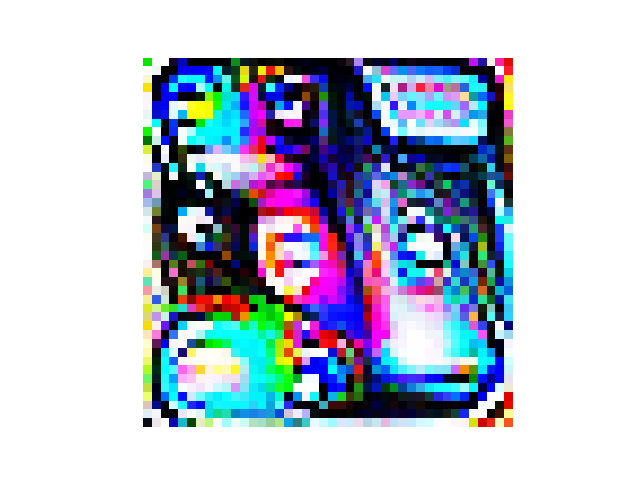}} &
        &
        \hbox{\includegraphics[scale=0.14]{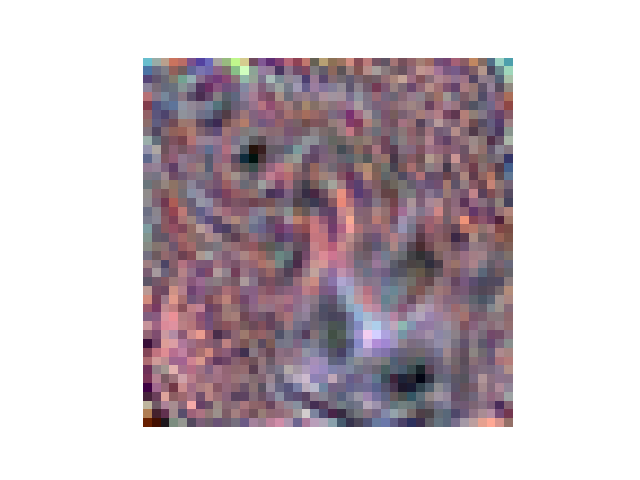}} &
        \hbox{\includegraphics[scale=0.14]{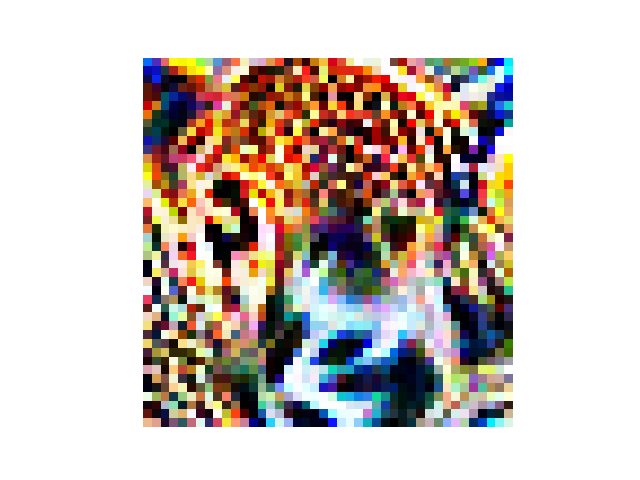}} \\
        \multicolumn{2}{c}{Baseball} & & \multicolumn{2}{c}{Cock} & &\multicolumn{2}{c}{iPod} & & \multicolumn{2}{c}{Jaguar}\\
        \hbox{\includegraphics[scale=0.14]{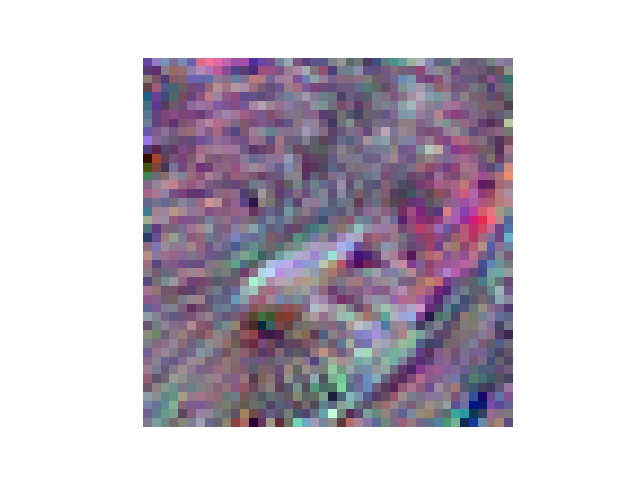}} &
        \hbox{\includegraphics[scale=0.14]{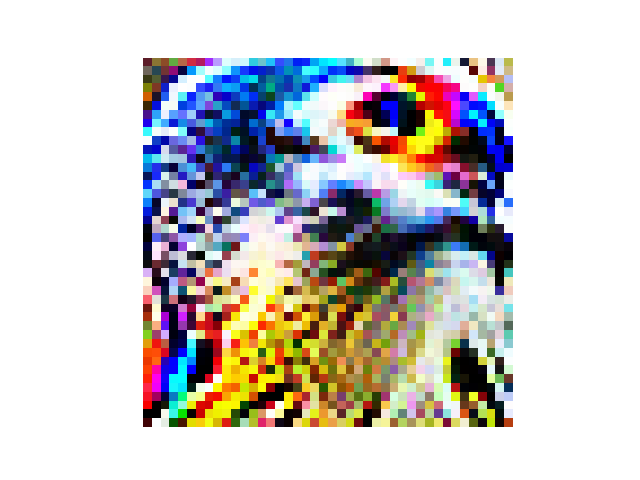}} &
        &
        \hbox{\includegraphics[scale=0.14]{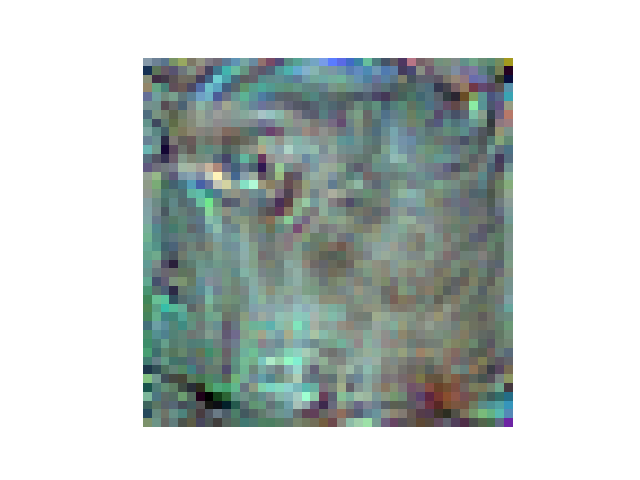}} &
        \hbox{\includegraphics[scale=0.14]{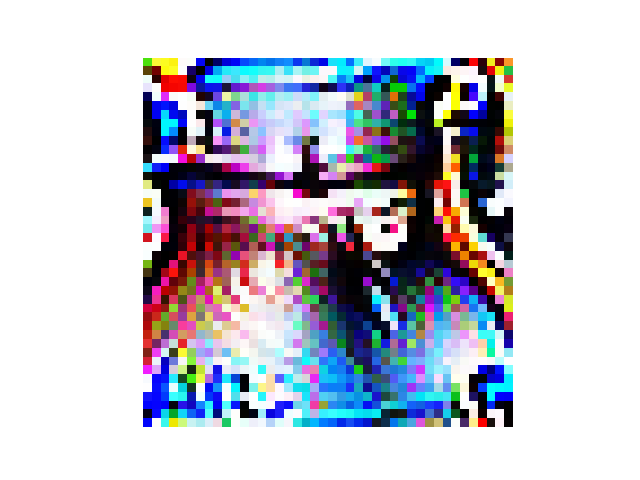}} &
        &
        \hbox{\includegraphics[scale=0.14]{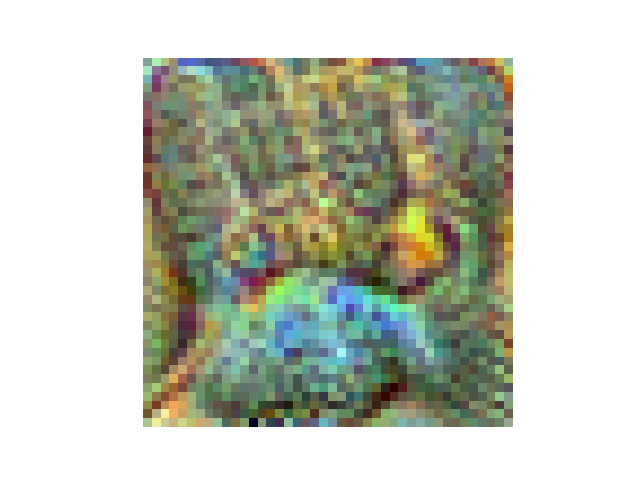}} &
        \hbox{\includegraphics[scale=0.14]{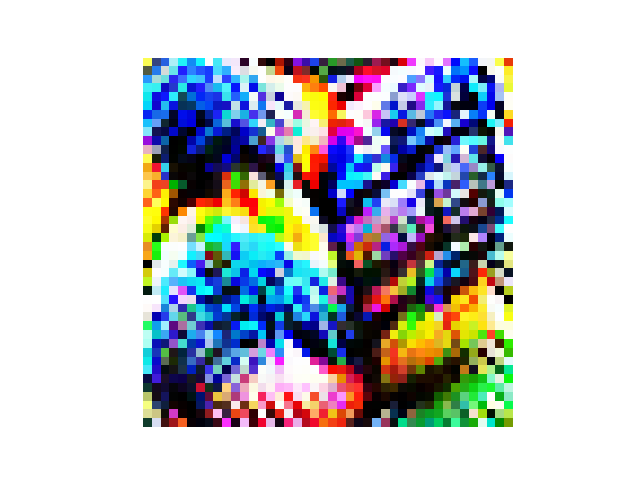}} &
        &
        \hbox{\includegraphics[scale=0.14]{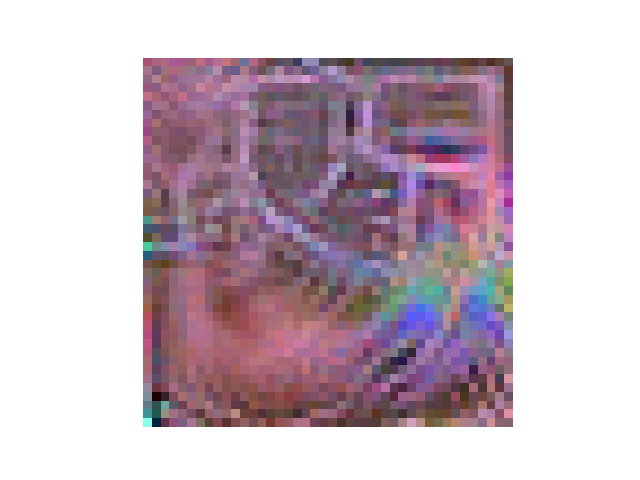}} &
        \hbox{\includegraphics[scale=0.14]{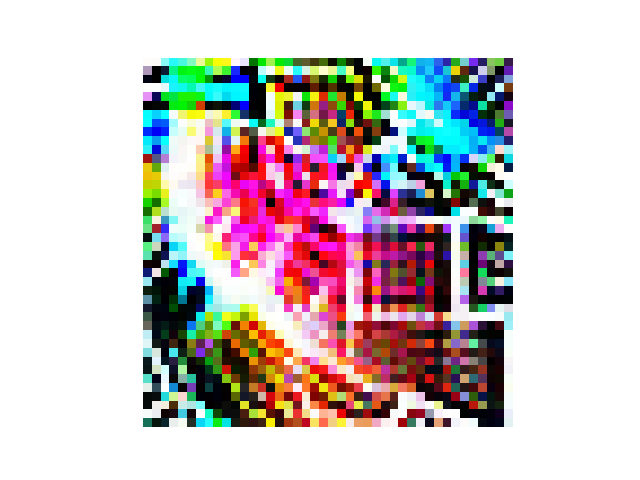}} \\
        \multicolumn{2}{c}{Kite} & & \multicolumn{2}{c}{Milk Can} & &\multicolumn{2}{c}{Mitten} & & \multicolumn{2}{c}{Monitor}\\
        \hbox{\includegraphics[scale=0.14]{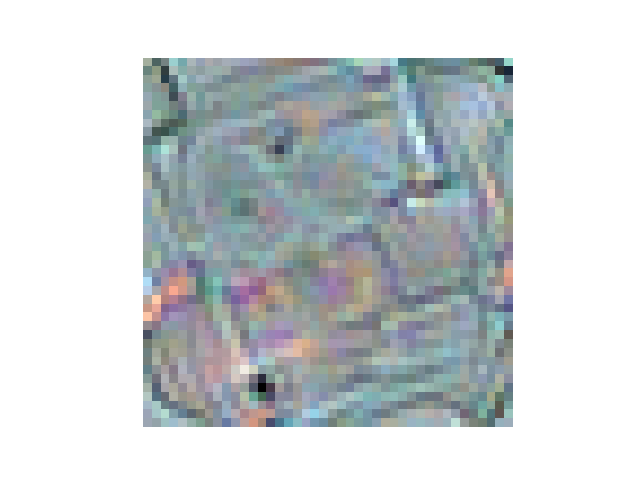}} &
        \hbox{\includegraphics[scale=0.14]{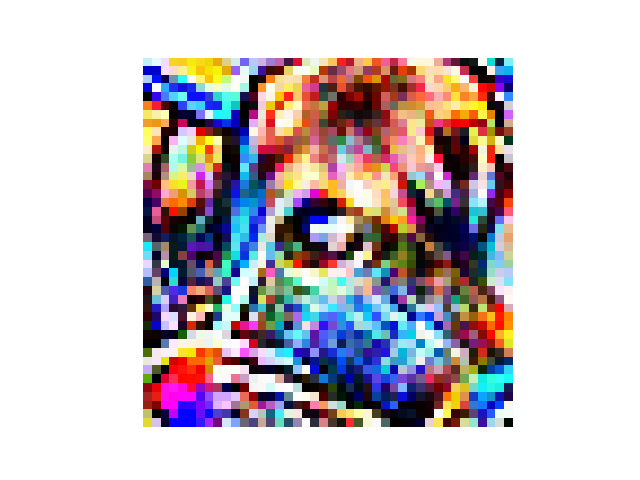}} &
        &
        \hbox{\includegraphics[scale=0.14]{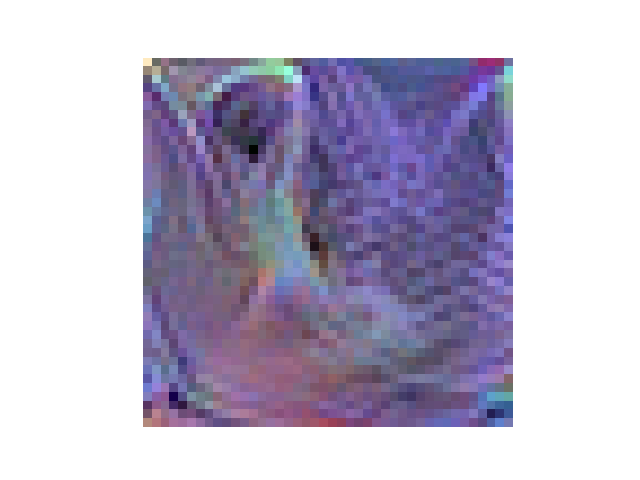}} &
        \hbox{\includegraphics[scale=0.14]{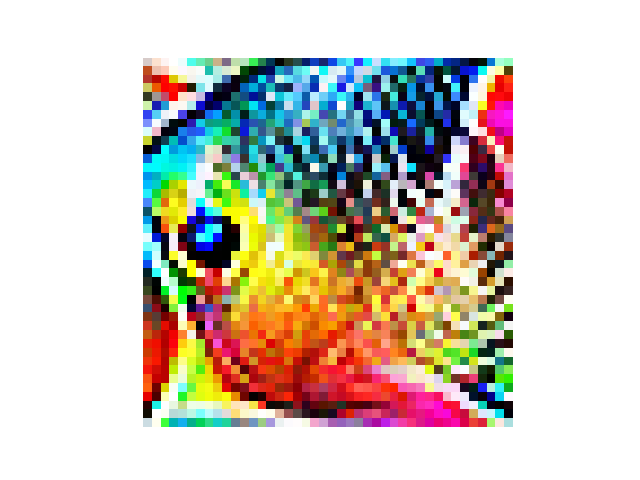}} &
        &
        \hbox{\includegraphics[scale=0.14]{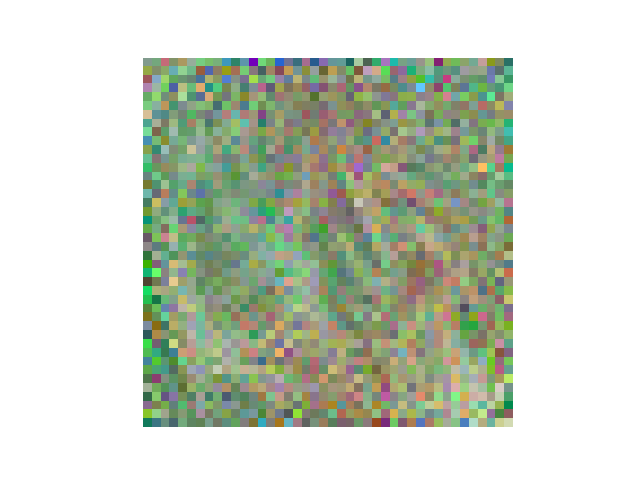}} &
        \hbox{\includegraphics[scale=0.14]{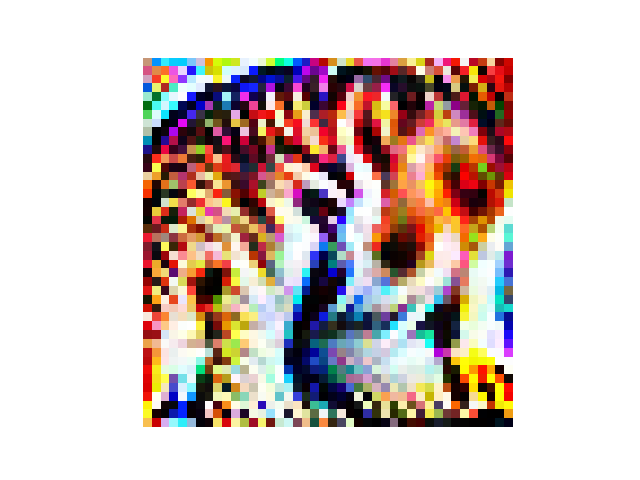}} &
        &
        \hbox{\includegraphics[scale=0.14]{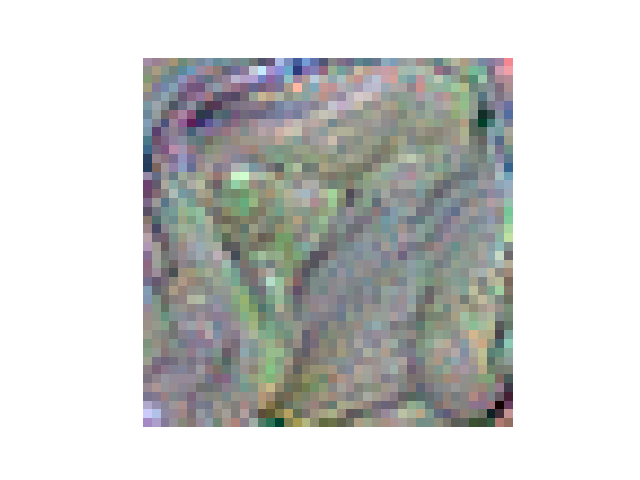}} &
        \hbox{\includegraphics[scale=0.14]{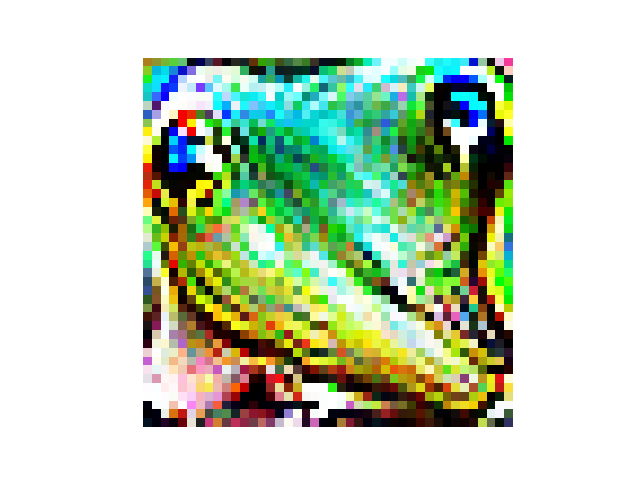}} \\
        \multicolumn{2}{c}{Muzzle} & & \multicolumn{2}{c}{Rock Beaut} & & \multicolumn{2}{c}{Rock Beauty} & & \multicolumn{2}{c}{Tree Frog}\\
        \hbox{\includegraphics[scale=0.14]{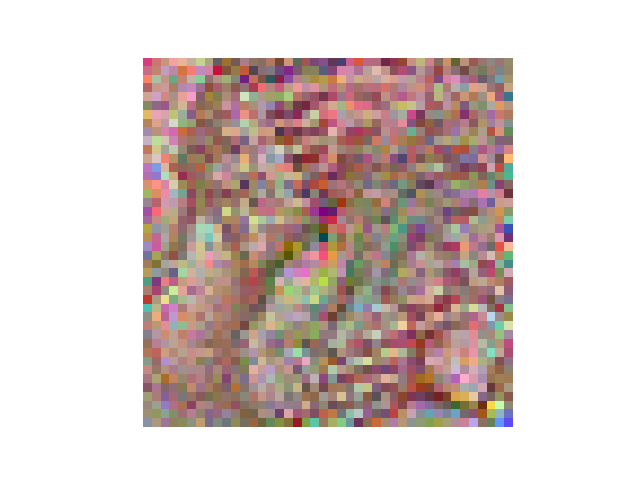}} &
        \hbox{\includegraphics[scale=0.14]{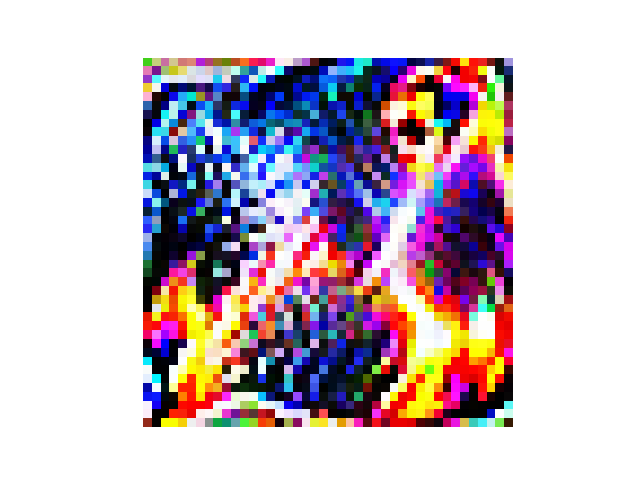}} &
        &
        \hbox{\includegraphics[scale=0.14]{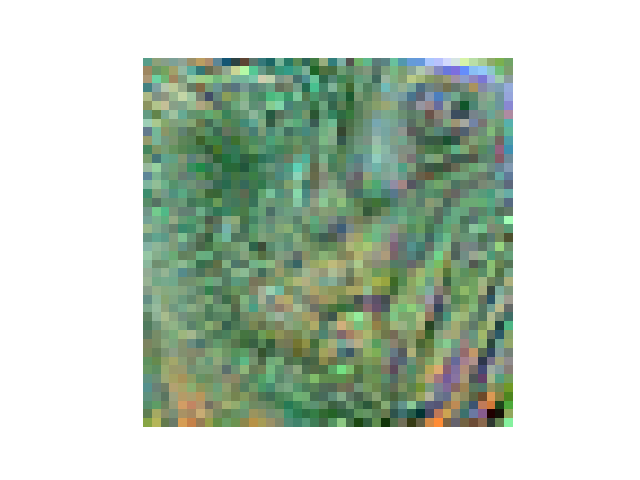}} &
        \hbox{\includegraphics[scale=0.14]{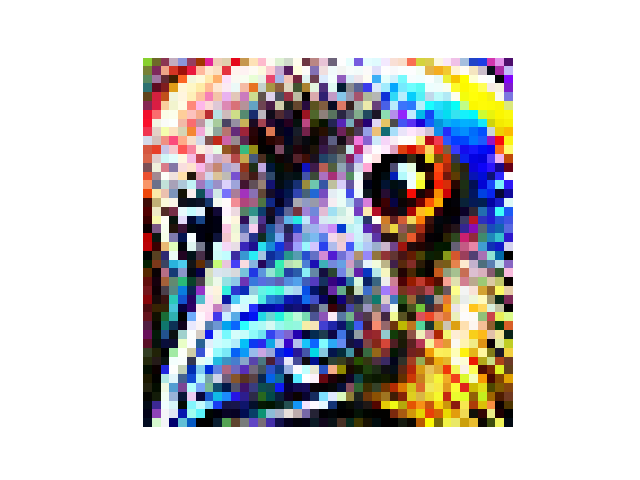}} \\
        \multicolumn{2}{c}{Volcano} & & \multicolumn{2}{c}{Vulture} \\
    \end{tabular}
    \caption{Network and image domain transferable localized noises for different target classes. Network domain noises were rescaled to image range for display purposes. Note that for many of the classes the resulting noises, and in particular the image-domain ones, contain prominent features of the target class like fur, eyes and global shape, and resembles a miniature version of the target class to some extent.}
        \label{fig:noises}
\end{figure*}

Overall we generated transferable noise patches for 14 different targets. The generated noises are presented in Figure \ref{fig:noises}, both for image domain noise patches and network domain patches (rescaled to image range for visualization purposes).

\paragraph{Network Domain}
We first provide anecdotal results of applying a trained noise to unseen images: Figure \ref{fig:transferable-noised} shows some examples of images with transferable noises in the network domain.

\begin{figure*}[t]
    \centering

    \tiny
    \begin{tabular}{cccc}
        \hbox{\includegraphics[scale=0.4]{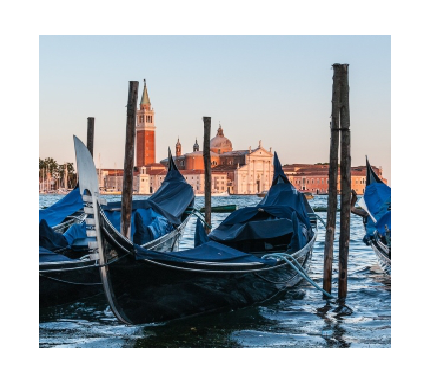}} &
        \hbox{\includegraphics[scale=0.4]{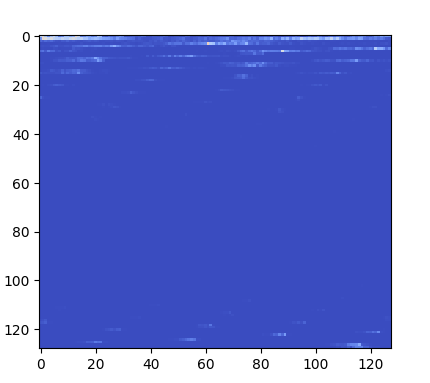}} &
        \hbox{\includegraphics[scale=0.4]{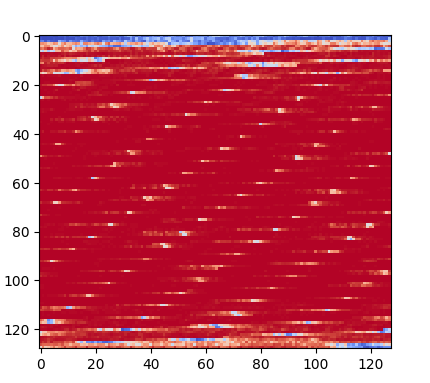}} &
        \hbox{\includegraphics[scale=0.4]{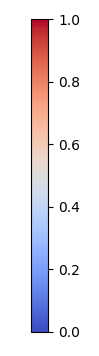}} \\
        Clean Image  & Predicted Probability for Source Class (Gondola) & Predicted Probability for Target Class (Volcano) & \\
        (Gondola) & Blue: 0, Red: 1 & Blue: 0, Red: 1 \\
        \hbox{\includegraphics[scale=0.4]{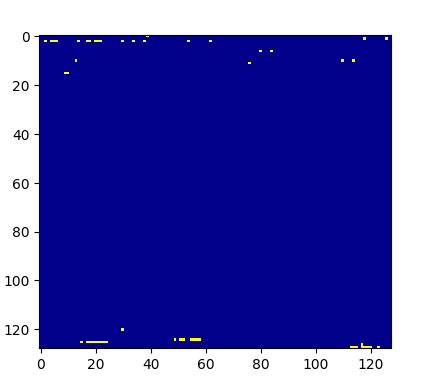}} &
        \hbox{\includegraphics[scale=0.4]{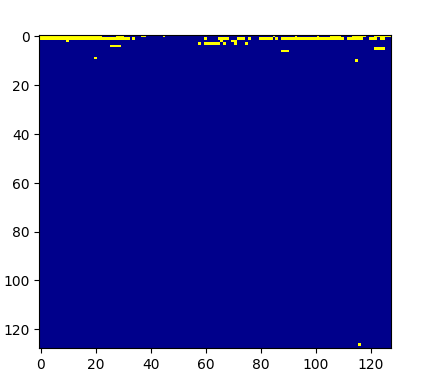}} &
        \hbox{\includegraphics[scale=0.4]{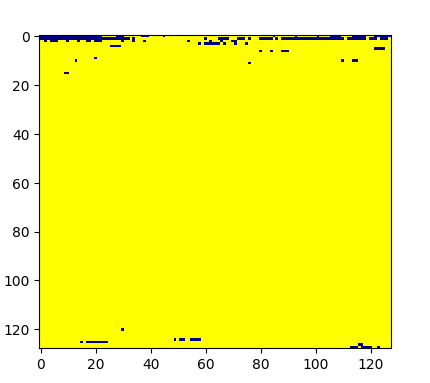}} & \\
        Yellow: Argmax is neither Gondola nor Volcano & Yellow: Argmax is Source Class (Gondola)  & Yellow: Argmax is Target Class (Volcano) &  \\
        Blue: Argmax either Gondola or Volcano & Blue: Argmax is some other class  & Blue: Argmax is some other class &  \\
    \end{tabular}
    \caption{Result of placing the Volcano network-domain noise patch on different locations in the Gondola image}
    \label{tbl:heatmap}
\end{figure*}

To assess the robustness to location, we applied a noise patch striding across every second pixel in an image that did not participate in the noise generation process, evaluating the model's prediction at each location using the following metrics: (1) probability of predicting the source class; (2) probability of predicting the target class; (3) argmax indication (i.e whether the class received the highest probability among other class) for the target class, source class, or neither classes. Figure \ref{tbl:heatmap} shows the result of this process for the (image:Gondola, target:Volcano) pair.\yg{I moved this to here.}

As can be seen, the noise is very robust, reducing the source class probability to near 0 across the image, and similarly increasing the target class probability.

We repeated this process for each of the 14 targets and each of the 100 images in our test set.
On average, applying the transferable noise to 83\% of the possible locations caused the target class to receive a score $\ge$ 0.9, and in 97\% of the locations it caused the model to not predict the original-source class. 

To evaluate the noise's transferability across images, we used a separate test set, consisting of 100 images from the ImageNet data-set that did not participate in the noise generation process. We placed each transferable noise patch on each image on the bottom right corner. Though only 43\% of our attempts made the model predict the target class with confidence $\ge$ 0.9, in 89\% of the cases it predicted the desired target as the most likely one, and in 100\% of the cases, the transferable noise patches prevented the model from predicting the original source class.

\paragraph{Image Domain} We evaluate the performance of the image domain noises using the same 100 images used in the network-domain evaluation. The success rates are somewhat lower, but still effective: while only 28.3\% of the attempts made the model predict the target class with confidence $\ge$ 0.9, in 74.1\% of the cases it caused the model to predict the target class as the most likely one, and in 78.9\% it resulted in a misclassification of the source class.
Figure \ref{fig:transferable_image_domain} shows some examples of transferable image-domain noises.

\begin{figure}[h]
 \centering
 \begin{tabular}{c}
 \includegraphics[scale=0.25]{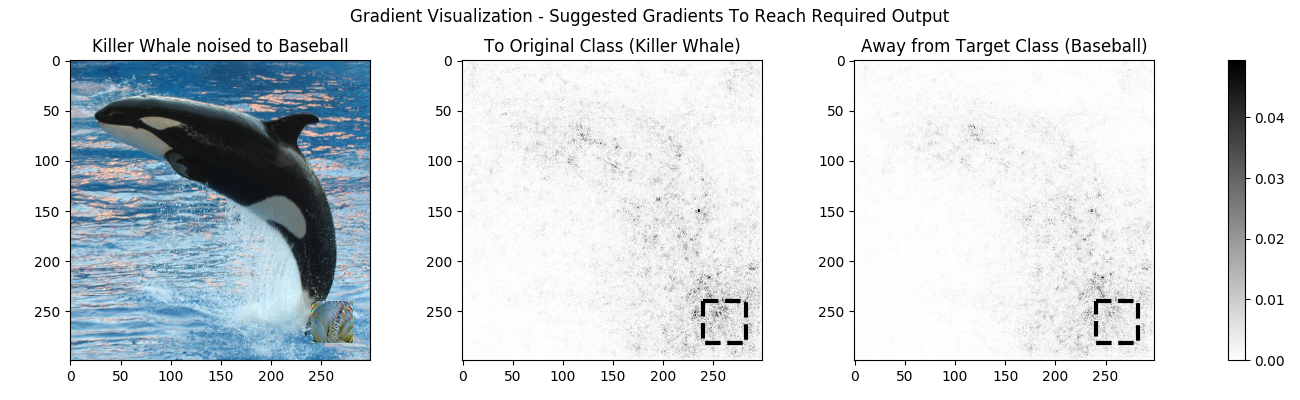}  \\
 \includegraphics[scale=0.25]{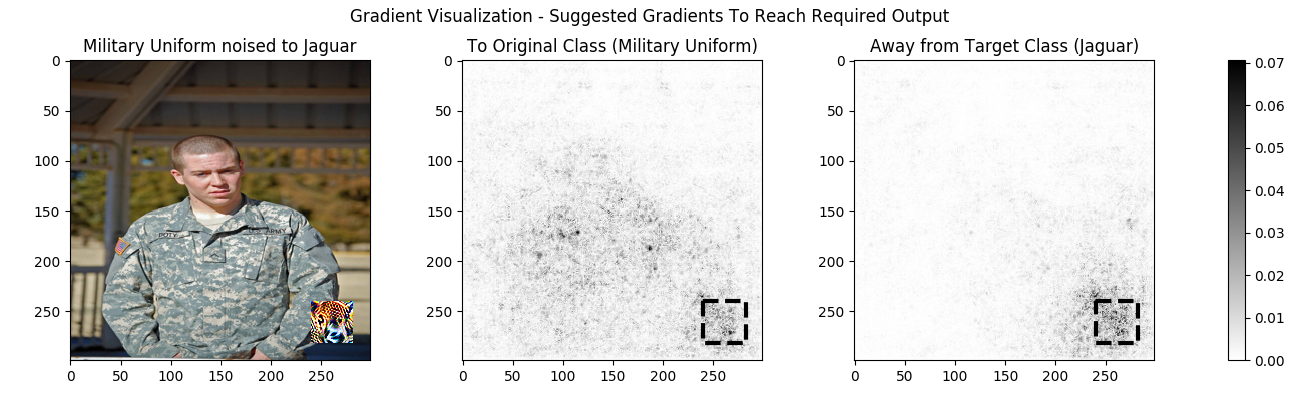}
 \end{tabular}
    \caption{Gradient updates required to fix the \texttt{Baseball} (86\%) (Top, Network Domain) and \texttt{Jaguar} (82\%) (Bottom, Image Domain) misclassifications back to \texttt{Dolphin} and \texttt{Uniform}. In both cases, the noised area in both cases is well represented in the gradients, but is not the most salient area.}
    \label{fig:gradients_fix_salient}
\end{figure}

\begin{figure}[h]
 \centering
 \begin{tabular}{c}

 \includegraphics[scale=0.25]{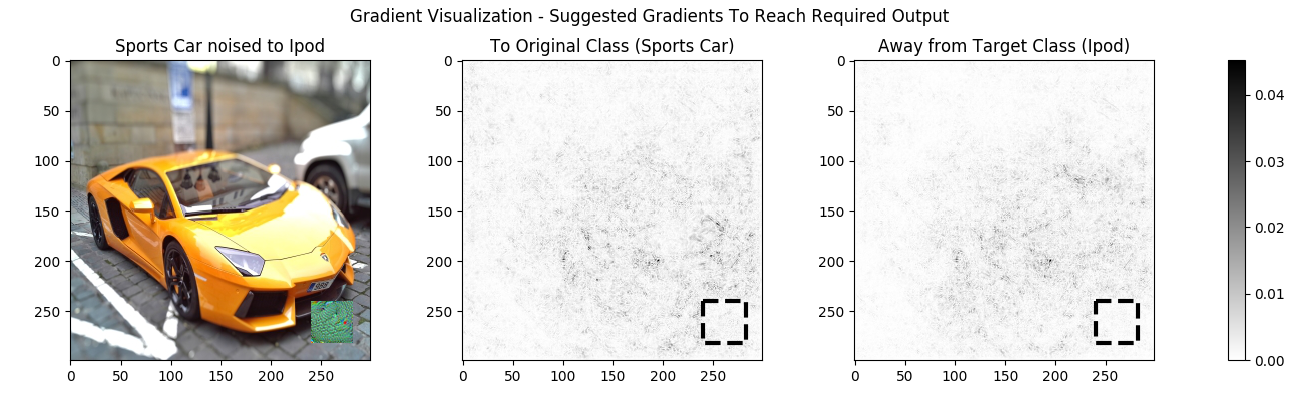} \\
 \includegraphics[scale=0.25]{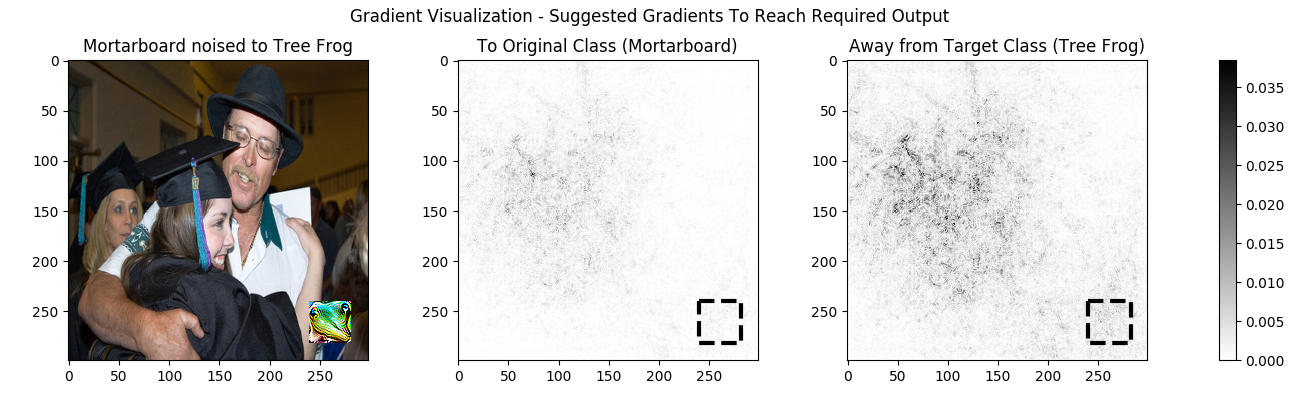}
 \end{tabular}
    \caption{Gradient updates required to fix the \texttt{iPod} (98\%) (Top, Network Domain) and the \texttt{Tree-frog} (80\%) (Bottom, Image Domain) misclassifications. Noised area in both cases is mostly ignored in the gradients.}
    \label{fig:gradients_fix_hidden}
\end{figure}

\paragraph{Class dependence}
\yg{moved this outside from "perceived by network". Can move back if desired.}
We now ask the following question: are specific classes better at serving as ``target'' classes, making it easier to fool the network when using them? Conversely we ask, are particular source classes harder such that attempting to apply noise patches them results in less success?

Figure \ref{fig:class_summary} summarizes the dependence of the noising process success on the source and target class using \textit{image domain} noises. Each cell shows the success rate of the noising process when using source images from ``source'' class and using ``target'' as the target. As can be seen there are significant differences between the different classes. Since we only cover a small subset of all possible classes in ImageNet we can only postulate that more globally structured classes such as ``volcano'' are more challenging to this localized process than locally structured classes as ''baseball''.

\Heatset{min=0,   
   max=1,   
   max colour=Burlywood2, 
   min colour=Snow1,      
   Min colour=OrangeRed1, 
   Max colour=SeaGreen3   
}

\begin{figure*}[h]
\tiny
 \centering
    \begin{tabular}{cc*{14}{H}ccccccccccccc}
     & \multicolumn{15}{c}{\textsf{Target Class}}\\
     & & \multicolumn1c{baseball}&\multicolumn1c{cock}&\multicolumn1c{iPod}&\multicolumn1c{jaguar}&\multicolumn1c{kite}&\multicolumn1c{milk-can}&\multicolumn1c{mitten}&\multicolumn1c{monitor}&\multicolumn1c{muzzle}&\multicolumn1c{rock beauty}&\multicolumn1c{tiger-cat}&\multicolumn1c{tree-frog}&\multicolumn1c{volcano}&\multicolumn1c{vulture}\\
 & academic gown 	&	 0.18 	&	 0.00 	&	 0.00 	&	 0.02 	&	 0.00 	&	 0.00 	&	 0.00 	&	 0.00 	&	 0.00 	&	 0.00 	&	 0.00 	&	 0.00 	&	 0.00 	&	 0.00 \\
 & banana 	&	 0.98 	&	 0.70 	&	 0.91 	&	 0.94 	&	 0.91 	&	 0.29 	&	 0.96 	&	 0.32 	&	 0.99 	&	 0.72 	&	 0.71 	&	 0.96 	&	 0.06 	&	 0.91 \\
 & barbershop 	&	 0.99 	&	 0.92 	&	 0.34 	&	 0.93 	&	 0.89 	&	 0.63 	&	 0.91 	&	 0.50 	&	 0.98 	&	 0.69 	&	 0.78 	&	 0.95 	&	 0.22 	&	 0.89 \\
 & bee 	&	 0.99 	&	 0.79 	&	 0.81 	&	 0.88 	&	 0.85 	&	 0.20 	&	 0.71 	&	 0.12 	&	 0.93 	&	 0.70 	&	 0.68 	&	 0.96 	&	 0.12 	&	 0.91 \\
 & bell pepper 	&	 0.94 	&	 0.48 	&	 0.19 	&	 0.73 	&	 0.66 	&	 0.50 	&	 0.53 	&	 0.04 	&	 0.12 	&	 0.53 	&	 0.61 	&	 0.89 	&	 0.05 	&	 0.90 \\
 & garden spider 	&	 0.97 	&	 0.66 	&	 0.10 	&	 0.95 	&	 0.89 	&	 0.46 	&	 0.19 	&	 0.22 	&	 0.99 	&	 0.84 	&	 0.83 	&	 0.96 	&	 0.05 	&	 0.71 \\
 & book jacket 	&	 0.99 	&	 0.77 	&	 0.15 	&	 0.81 	&	 0.90 	&	 0.83 	&	 0.45 	&	 0.27 	&	 0.99 	&	 0.90 	&	 0.84 	&	 0.97 	&	 0.08 	&	 0.96 \\
 & brown bear 	&	 0.80 	&	 0.21 	&	 0.13 	&	 0.30 	&	 0.32 	&	 0.31 	&	 0.21 	&	 0.02 	&	 0.50 	&	 0.19 	&	 0.32 	&	 0.55 	&	 0.01 	&	 0.28 \\
& bullet train 	&	 0.98 	&	 0.47 	&	 0.01 	&	 0.79 	&	 0.61 	&	 0.54 	&	 0.65 	&	 0.20 	&	 0.69 	&	 0.40 	&	 0.74 	&	 0.88 	&	 0.05 	&	 0.49 \\
& cardigan 	&	 0.94 	&	 0.51 	&	 0.25 	&	 0.77 	&	 0.66 	&	 0.18 	&	 0.55 	&	 0.17 	&	 0.93 	&	 0.29 	&	 0.56 	&	 0.71 	&	 0.15 	&	 0.45 \\
& corn 	&	 0.99 	&	 0.76 	&	 0.63 	&	 0.96 	&	 0.97 	&	 0.89 	&	 0.90 	&	 0.05 	&	 1.00 	&	 0.81 	&	 0.72 	&	 0.95 	&	 0.33 	&	 0.98 \\
& crash helmet 	&	 0.98 	&	 0.50 	&	 0.10 	&	 0.79 	&	 0.80 	&	 0.22 	&	 0.41 	&	 0.05 	&	 0.58 	&	 0.46 	&	 0.58 	&	 0.92 	&	 0.02 	&	 0.41 \\
& daisy 	&	 0.96 	&	 0.51 	&	 0.39 	&	 0.77 	&	 0.73 	&	 0.71 	&	 0.66 	&	 0.33 	&	 0.87 	&	 0.62 	&	 0.65 	&	 0.90 	&	 0.13 	&	 0.72 \\
& dishrag 	&	 0.97 	&	 0.67 	&	 0.06 	&	 0.90 	&	 0.95 	&	 0.52 	&	 0.63 	&	 0.02 	&	 0.95 	&	 0.67 	&	 0.73 	&	 0.92 	&	 0.10 	&	 0.92 \\
& dogsled 	&	 0.93 	&	 0.44 	&	 0.14 	&	 0.48 	&	 0.62 	&	 0.60 	&	 0.25 	&	 0.02 	&	 0.49 	&	 0.27 	&	 0.44 	&	 0.70 	&	 0.07 	&	 0.32 \\
& electric ray 	&	 0.98 	&	 0.77 	&	 0.15 	&	 0.92 	&	 0.94 	&	 0.98 	&	 0.89 	&	 0.14 	&	 0.99 	&	 0.84 	&	 0.73 	&	 0.95 	&	 0.13 	&	 0.98 \\
& football helmet 	&	 0.96 	&	 0.61 	&	 0.01 	&	 0.65 	&	 0.77 	&	 0.00 	&	 0.05 	&	 0.01 	&	 0.16 	&	 0.41 	&	 0.50 	&	 0.84 	&	 0.00 	&	 0.66 \\
& gong 	&	 0.95 	&	 0.36 	&	 0.03 	&	 0.84 	&	 0.71 	&	 0.11 	&	 0.42 	&	 0.02 	&	 0.50 	&	 0.28 	&	 0.72 	&	 0.75 	&	 0.01 	&	 0.37 \\
& green mamba 	&	 0.92 	&	 0.23 	&	 0.11 	&	 0.56 	&	 0.55 	&	 0.35 	&	 0.25 	&	 0.10 	&	 0.81 	&	 0.35 	&	 0.53 	&	 0.43 	&	 0.04 	&	 0.36 \\
& hatchet 	&	 0.50 	&	 0.00 	&	 0.00 	&	 0.00 	&	 0.01 	&	 0.00 	&	 0.00 	&	 0.00 	&	 0.01 	&	 0.00 	&	 0.00 	&	 0.07 	&	 0.00 	&	 0.00 \\
& hip 	&	 0.98 	&	 0.65 	&	 0.41 	&	 0.95 	&	 0.82 	&	 0.95 	&	 0.96 	&	 0.29 	&	 0.99 	&	 0.80 	&	 0.79 	&	 0.83 	&	 0.20 	&	 0.59 \\
\rotatebox{90}{\makebox[0pt]{\textsf{Source Class}}}
& hummingbird 	&	 1.00 	&	 0.61 	&	 0.66 	&	 0.94 	&	 0.91 	&	 0.91 	&	 0.69 	&	 0.57 	&	 0.96 	&	 0.92 	&	 0.86 	&	 0.97 	&	 0.19 	&	 0.94 \\
& Indian cobra 	&	 0.96 	&	 0.29 	&	 0.30 	&	 0.76 	&	 0.64 	&	 0.36 	&	 0.14 	&	 0.16 	&	 0.87 	&	 0.60 	&	 0.59 	&	 0.77 	&	 0.05 	&	 0.64 \\
& iPod 	&	 0.99 	&	 0.79 	&	 1.00 	&	 0.96 	&	 0.88 	&	 0.10 	&	 0.45 	&	 0.10 	&	 0.94 	&	 0.80 	&	 0.89 	&	 0.93 	&	 0.13 	&	 0.94 \\
& jellyfish 	&	 0.99 	&	 0.52 	&	 0.75 	&	 0.91 	&	 0.87 	&	 0.93 	&	 0.56 	&	 0.31 	&	 0.96 	&	 0.84 	&	 0.89 	&	 0.94 	&	 0.05 	&	 0.89 \\
& lacewing 	&	 0.93 	&	 0.33 	&	 0.05 	&	 0.72 	&	 0.73 	&	 0.20 	&	 0.24 	&	 0.02 	&	 0.90 	&	 0.17 	&	 0.36 	&	 0.46 	&	 0.05 	&	 0.15 \\
& mailbag 	&	 0.84 	&	 0.23 	&	 0.02 	&	 0.44 	&	 0.60 	&	 0.13 	&	 0.10 	&	 0.00 	&	 0.69 	&	 0.02 	&	 0.68 	&	 0.73 	&	 0.02 	&	 0.05 \\
& mantis 	&	 0.94 	&	 0.47 	&	 0.27 	&	 0.71 	&	 0.66 	&	 0.27 	&	 0.35 	&	 0.19 	&	 0.59 	&	 0.39 	&	 0.67 	&	 0.73 	&	 0.10 	&	 0.59 \\
& military uniform 	&	 0.97 	&	 0.82 	&	 0.15 	&	 0.90 	&	 0.87 	&	 0.69 	&	 0.90 	&	 0.33 	&	 0.97 	&	 0.68 	&	 0.67 	&	 0.90 	&	 0.21 	&	 0.82 \\
& minivan 	&	 0.98 	&	 0.87 	&	 0.13 	&	 0.92 	&	 0.91 	&	 0.76 	&	 0.95 	&	 0.09 	&	 1.00 	&	 0.78 	&	 0.80 	&	 0.97 	&	 0.09 	&	 0.92 \\
& mortarboard 	&	 0.97 	&	 0.48 	&	 0.24 	&	 0.60 	&	 0.66 	&	 0.10 	&	 0.55 	&	 0.07 	&	 0.66 	&	 0.36 	&	 0.51 	&	 0.67 	&	 0.06 	&	 0.56 \\
& poncho 	&	 0.99 	&	 0.89 	&	 0.51 	&	 0.92 	&	 0.93 	&	 0.37 	&	 0.93 	&	 0.18 	&	 1.00 	&	 0.78 	&	 0.88 	&	 0.95 	&	 0.05 	&	 0.92 \\
& rapeseed 	&	 0.99 	&	 0.71 	&	 0.51 	&	 0.93 	&	 0.90 	&	 0.91 	&	 0.96 	&	 0.27 	&	 1.00 	&	 0.86 	&	 0.75 	&	 0.94 	&	 0.38 	&	 0.83 \\
& starfish 	&	 0.99 	&	 0.82 	&	 0.69 	&	 0.95 	&	 0.77 	&	 0.50 	&	 0.86 	&	 0.17 	&	 0.99 	&	 0.82 	&	 0.74 	&	 0.98 	&	 0.07 	&	 0.95 \\
& turnstile 	&	 0.99 	&	 0.78 	&	 0.01 	&	 0.93 	&	 0.74 	&	 0.50 	&	 0.56 	&	 0.02 	&	 0.96 	&	 0.23 	&	 0.72 	&	 0.94 	&	 0.05 	&	 0.73 \\
& velvet 	&	 0.98 	&	 0.66 	&	 0.19 	&	 0.91 	&	 0.90 	&	 0.61 	&	 0.86 	&	 0.04 	&	 1.00 	&	 0.63 	&	 0.49 	&	 0.92 	&	 0.11 	&	 0.82 \\
& walking stick 	&	 0.97 	&	 0.40 	&	 0.54 	&	 0.79 	&	 0.78 	&	 0.54 	&	 0.54 	&	 0.27 	&	 0.67 	&	 0.59 	&	 0.76 	&	 0.84 	&	 0.10 	&	 0.69 \\
& washer 	&	 0.98 	&	 0.64 	&	 0.17 	&	 0.90 	&	 0.78 	&	 0.51 	&	 0.78 	&	 0.25 	&	 0.93 	&	 0.61 	&	 0.80 	&	 0.93 	&	 0.07 	&	 0.66 \\
& water snake 	&	 0.99 	&	 0.75 	&	 0.59 	&	 0.94 	&	 0.97 	&	 0.94 	&	 0.86 	&	 0.22 	&	 1.00 	&	 0.89 	&	 0.69 	&	 0.95 	&	 0.28 	&	 0.96 \\
& wing 	&	 0.99 	&	 0.81 	&	 0.50 	&	 0.93 	&	 0.99 	&	 0.96 	&	 0.98 	&	 0.27 	&	 0.96 	&	 0.94 	&	 0.74 	&	 0.93 	&	 0.49 	&	 0.99 \\
    \end{tabular}
 \caption{Average target score heatmap for transferable image-domain noises. The value in $C_{i,j}$ is the average score assigned to the target class $j$ when that the source class is $i$.}
  \label{fig:class_summary}
\end{figure*}

\section{Noise perceived by the network}

We successfully generated small transferable, localized noise patches that fool a state-of-the-art network into misclassifying an example. 

In \cite{brown2017adversarial}, the authors suggest that the localized noise works because it is much more salient than natural looking objects in the scene, capturing all of the network's attention. To try and examine this claim, we take some noised images and attempt to ``fix'' the effect of the noise on the classification output. 

First, we attempt to fix a noised image by feeding it into the network and taking gradient steps over the entire image towards the original source class.\footnote{This is similar to the salience map computation in \cite{simonyanVZ13}, but we accumulate the absolute values of several gradient steps, until reaching the desired outcome.}
For example, 
we take a noised image of a \texttt{Killer Whale} classified as a \texttt{Baseball}, and take gradient steps over the entire image until it is classified as a \texttt{Killer Whale}. The top image triplet in Figure \ref{fig:gradients_fix_salient} visualizes the gradient fix for the whale image. The center image shows the absolute values of the cumulative updates that were needed to overcome the noise and re-classify as a Whale. Notice that while a lot of the gradient efforts are within the noise patch, there is also a lot of activity \emph{outside} of the patch, accenting features of the Whale's body.

What if instead of attempting to optimize towards the source class, we optimize the noised image \emph{away from} the target class? I.e., rather than pushing the noised image to ``be a whale again'', we push it to ``stop being a baseball''? The resulting cumulative gradients are displayed on the top images in Figure \ref{fig:gradients_fix_salient} (whale image triplet). We see less gradient activity overall, and, somewhat surprisingly, there again seems to be at least as much activity outside the noise patch as it is within it. This gradient behavior persists on different image/noise cases, with a combination of gradient within the noise box and outside of is. 

Some noise cases are more intriguing. For example, the top image triplet in
Figure \ref{fig:gradients_fix_hidden} shows the case of a \texttt{Sports Car} being noised to a \texttt{iPod} at 98\%. 
When driving the image towards \texttt{Sports Car} (center), the gradient seem to accent the car's front and its surrounding, while \emph{almost completely ignoring} the noised patch. 
In order to restore the original class, the network's gradients work to enhance features of the original class rather than diminishing the intruding noise.
This pattern persists when optimizing the noised image \emph{away from} \texttt{iPod} (right): there are hardly any updates within the noised area. The stealthiness of the noise is observed both in network-domain noises (Fig \ref{fig:gradients_fix_hidden} top) and in image-domain noises (Fig \ref{fig:gradients_fix_hidden} bottom).

This suggests that, at least in some image/noise combinations, the localized visible noise is not only very effective at misleading the classifier and triggering the target class, it is also quite \emph{stealthy}--at least according to the target gradients wrt to the image, the classifier attributes the target to various elements in the image, but not to the noisy patch itself. This is contrast to previous suggestions that noise is somehow more salient to the network than other parts of the image.

\subsection{Quantifying the saliency of noise}
\yg{I rewrote this part. Daniel, please verify that it is understandable, and fix if not.}
To further quantify the above behavior, we consider the gradients required to fix the 2,800 noised images that result from applying each of the 28 noises in Figure \ref{fig:noises} to the bottom right corner of each of the 100 test images. 

In each gradient-fix map, we score each 42x42 patch according to: (1) the maximum absolute value within the patch (MAX); and (2) the sum of the absolute values within the patch (SUM). We then take the patch with the maximum value according to each metric, and check if it overlaps with the noise location. Table \ref{tbl:quantify_fix} lists the results. According to both metrics, the most active patch very rarely overlaps with the noise location. This strengthens our previous claim and suggests that the model may fail to understand and locate the noise in the image.

\begin{table}[th]
\centering
\begin{tabular}{l|l|c|c}
\hline
Domain & Fix method    & MAX  & SUM \\ \hline \hline
Network & Towards Source   & 0.4\%     & 0.15\%  \\ 
Network & Away from Target & 0.5\%     & 0.13\%  \\ \hline
Image & Towards Source   & 0.6\%      & 5.4\%  \\ 
Image & Away from Target & 0.7\& & 5.2\%      \\ \hline
\end{tabular}
\caption{Quantifying the saliency of transferable noises.
Each cell lists the percent of out of 1,400 (images,target) pairs in which the most active 42x42 gradient-fix patch ovelaps with the 42x42 noise patch.
MAX: active patches are selected according to the max value in the patch. SUM: selected according to sum of values in the patch.}
\label{tbl:quantify_fix}
\end{table}

\section{Related Work}
\label{sec:related}

Most works focus on adding a small amount of noise, imperceptible to the human eye, but covering the entire image. It was shown that it is possible to construct such noises to thwart multiclass image classifiers \cite{goodfellow2014explaining,szegedy2014intriguing,dezfooli2016universal}, and, more recently, also structured classifiers in tasks such as image segmentation and pose estimation \cite{cisse2017houdini}.
In all these cases the noise covers the entire image, including the salient objects within it. We focus on cases where the noise may be larger, but only affect a small part of the image.

Several works explore setups that do not noise the entire image.
Sharif et al \cite{sharif2016accessorize} causes a face recognition system to identify an incorrect face by adding glasses with a specially constructed frame texture, and Evtimov et al \cite{evtimov2017robust} causes misclassification of traffic signs by adding a specific pattern of rectangular, solid-colored patches on top of a traffic sign. These are very impressive works in which the attack transfers to images taken in the real world. However, from the noise locality perspective, in both these cases, the noise patches attack very prominent characteristics of the objects to be classified. The glasses hide areas around the eyes, which are very indicative for face recognition. Similarly, many traffic signs can be abstracted as a specific arrangement of solid-colored rectangles on a solid-colored background, so it is not too surprising that messing with the rectangles formation can fool a traffic-sign classifier. 
Su et al \cite{su2017one} demonstrate that, in about 70\% of 32x32 images from CIFAR-10, it is possible to find a single pixel whose change in value can cause a misclassification (the number drops to between 20\% and 30\% for a targeted attack towards a specific target class). However, this requires a very small image size, and the offending pixel usually at the center of the classified image. Papernot et al \cite{papernot2016limitations} show that, when considering black-and-white digits classification, changing a small percentage of pixels from black to white can cause misclassification. The pixels are spread out across the entire image and look as harming noise to a human observer. 

In contrast to all these works, in our case, the pixels are localized to a specific region of the image and do not cover the source-class object at all.  Finally, the recent \emph{adversarial patch} work by Brown et al  \cite{brown2017adversarial} is very similar to our setting, and was discussed throughout this report. They focus on printable patches that are immune to scaling or ration, and that can be used for physical attacks. As a consequence, their patches are required to be very large, covering upward from 10\% of the image in order to be effective. We demonstrate that, when relaxing these requirements, networks can be fooled also by much smaller patches of visible noise, that cover a substantially smaller area of the image. Moreover, we show that---at least according to its gradients---the network does not recognize the attacking patches as the source of the misclassification.

\section{Conclusions}
We show that it is possible to learn visible and localized adversarial noise patches that cover only 2\% of the pixels in the image, none of them on the main salient object, and that cause a state-of-the-art image classifier to misclassify to arbitrary labels. Such noise patches are not image specific, and the same patch can be applied to arbitrary images and locations, causing a misclassification to the desired target class with very high success rates. Perhaps surprisingly, in many cases, the noise patch is not particularly salient as far as the network is concerned as revealed by the resulting gradients when one attempts to ``fix'' the noise effect. 

The resulting noise patches resemble the target classes both in the network domain and image domain cases. Prominent features such as texture and fur, body parts like eyes and beaks and global shape characteristics are all reflected in the outputs. The noise patch---however similar to the target class---is still much smaller than main object in the image, yet makes the network misclassify. 

Furthermore, and somewhat surprisingly when considering that the input patch lies pretty close to stimuli the network usually observes (certainly in the image domain case), the gradients when attempting to revert the noise tell us that the noised patch is not very salient to the network when compared to other locations within the image.

Considering the noised patch is not salient to the network raises some concerns from a security perspective, but also raises some interesting questions about the inner workings of the network. From a security perspective, such a noise patch is problematic as it seems that the network does not ``see'' that is being fooled, possibly making it hard protect against such attacks. This is true both for the image domain and network domain noise. From a scientific perspective we feel the methods and results presented here may serve as the foundation of some interesting future research. The brittleness of the classification output of these networks tells us something about the way they operate. One could ask whether different architecture are susceptible to such attacks in different ways and whether specific architectural choices affect the resulting vulnerabilities. Furthermore, it raises the question whether \emph{during training} we can try to make these systems more robust, incorporating adversarial auxiliary losses into the training pipeline and making the networks more robust. These are all left for future research.


\end{document}